\documentclass[10pt,letterpaper]{article}
\usepackage[a4paper,
            bindingoffset=0.2in,
            left=1in,
            right=1in,
            top=1in,
            bottom=1in,
            footskip=.25in]{geometry}
\usepackage{tabularx,booktabs}
\usepackage{times}
\usepackage{hyperref}
\usepackage{supertabular}
\usepackage{url}
\usepackage{fullpage}
\usepackage{graphicx}
\usepackage{lipsum}
\usepackage{algorithm}
\usepackage[noend]{algorithmic}
\usepackage{appendix}
\usepackage{authblk}
\usepackage{multirow}
\usepackage{multicol}
\usepackage{bigstrut}

\newcommand{\junk}[1]{}

\begin{document}
\thispagestyle{empty}

\title{CLMB: deep contrastive learning for robust metagenomic binning}


\author[1,2]{Pengfei Zhang}
\author[1,2]{Zhengyuan Jiang}
\author[1,4]{Yixuan Wang}
\author[1,3]{Yu Li \thanks{Corresponding Author. Email: liyu@cse.cuhk.edu.hk}}
\affil[1]{\small Department of Computer Science and Engineering, CUHK, Hong Kong SAR, China}
\affil[2]{\small University of Science and Technology of China, Hefei, Anhui, China}
\affil[3]{\small The CUHK Shenzhen Research Institute, Hi-Tech Park, Nanshan, Shenzhen, 518057, China}
\affil[4]{\small Department of Mathematics, HIT, 264209 Weihai, China}

\date{}

\maketitle
\begin{abstract}
The reconstruction of microbial genomes from large metagenomic datasets is a critical procedure for finding uncultivated microbial populations and defining their microbial functional roles.
To achieve that, we need to perform metagenomic binning, clustering the assembled contigs into draft genomes. Despite the existing computational tools, most of them neglect one important property of the metagenomic data, that is, the noise. To further improve the metagenomic binning step and reconstruct better metagenomes, we propose a deep Contrastive Learning framework for Metagenome Binning (CLMB), which can efficiently eliminate the disturbance of noise and produce more stable and robust results. 
Essentially, instead of denoising the data explicitly, we add simulated noise to the training data and force the deep learning model to produce similar and stable representations for both the noise-free data and the distorted data. Consequently, the trained model will be robust to noise and handle it implicitly during usage.
CLMB outperforms the previous state-of-the-art binning methods significantly, recovering the most near-complete genomes on almost all the benchmarking datasets (up to 17\% more reconstructed genomes compared to the second-best method).
It also improves the performance of bin refinement, reconstructing 8-22 more high-quality genomes and 15-32 more middle-quality genomes than the second-best result. Impressively, in addition to being compatible with the binning refiner, single CLMB even recovers on average 15 more HQ genomes than the refiner of VAMB and Maxbin on the benchmarking datasets.
On a real mother-infant microbiome dataset with 110 samples, CLMB is scalable and practical to recover 365 high-quality and middle-quality genomes (including 21 new ones), providing insights into the microbiome transmission.
CLMB is open-source and available at  \href{https://github.com/zpf0117b/CLMB/}{https://github.com/zpf0117b/CLMB/}.





\end{abstract}

\newpage

\section{Introduction}
Studies of microbial communities are increasingly dependent on high-throughput, whole-genome shotgun sequencing datasets \cite{vajt14,tr05}.
General studies assemble short sequence reads obtained from metagenome sequencing into longer sequence fragments (contigs), and subsequently group them into genomes by metagenome binning \cite{qws17,mks10}. 
Metagenome binning is a crucial step in recovering the genomes, which therefore provides access to uncultivated microbial populations and understanding their microbial functional roles.


In recent years, we have witnessed great progress in metagenome binning.
Firstly, the composition and the abundance of each contig are proved useful for binning \cite{abd14,kbd09}.
Secondly, several programs have been developed for fully automated binning procedures, which leverage both composition and abundance as features.
MetaBAT \cite{kfem15}, MetaBAT2 \cite{klkteaw19}, CONCOCT \cite{abd14}, and Maxbin2 \cite{wss16} utilize the composition and abundance information and take the metagenome binning as the clustering task.
VAMB \cite{nja21} performs dimensionality reduction, encoding the data using VAE first and subsequently conducting the clustering task.
Thirdly, a new approach ‘multi-split’ is developed and achieves great performance \cite{nja21,zbpz21}. It gathers contigs from all the samples and calculates the abundance among samples, clustering them into bins and splitting the bins by sample.

Earlier works on metagenomics binning achieved good performance by applying different strategies for clustering.
However, they ignored the potential factors in real-world conditions that influence the quality of metagenomic short reads, such as the low total biomass of microbial-derived genomes in clinical isolates \cite{vbr20} and the imperfect genomic sequencing process, for example, base substitutions, insertions, and deletions \cite{fow19}. 
As a consequence of the factors, metagenomic sequences are susceptible to the noise issue, such as contamination noise and alignment noise \cite{vbr20}.
The potential noise can influence the quality of metagenomics sequences, and therefore make it difficult to distinguish whether certain contigs come from the same type of or different bacterial genomes, impacting the correctness of the formed draft genomes.
Furthermore, all of the existing binners are restricted by data volume. 

To learn a high-quality draft genome for each bacterium, we design a novel deep Contrastive Learning algorithm for Metagenomic Binning (CLMB) to handle the noise (Figure \ref{overview}). The basic idea of the CLMB module is that, since the noise of the real dataset is hard to detect, we add simulated noise to the data and force the trained model to be robust to them. 
Essentially, instead of denoising the data explicitly, we add simulated noise to the training data and ask the deep learning model to produce similar and stable representations for both the noise-free data and the distorted data. Consequently, the trained model will be robust to noise and handle it implicitly during usage.
By effectively tackling the noise in the metagenomics data using the contrastive deep learning framework \cite{hk20,han2021self}, we can group pairs of contigs that originate from the same type of bacteria together while dividing contigs from different species to different bins.
Moreover, CLMB performs data augmentation before training and take the augmented data as training data. Unlike other binners, CLMB uses the augmented data, instead of the raw data, for training. Therefore, 
the data volume for training is largely increased, which improves the representation of the deep learning model and prevents overfitting.
CLMB also keeps the ‘multi-split’ approach, which combines the contigs of all the samples for binning, because the contrastive deep learning benefits more from a larger data size \cite{hk20}.

On the CAMI2 Toy Human Microbiome Project Dataset\cite{shb17}, CLMB outperforms the previous state-of-the-art binning methods significantly, recovering the most near-complete genomes on almost all the benchmarking datasets. Specifically, CLMB reconstructs up to 17\% more near-complete genomes compared to the second-best method. We then investigate the recovered genomes under different criteria and find that more information contained in data contributes to the binning performance of CLMB. By involving CLMB, the performance of bin refinement is improved, reconstructing 8-22 more high-quality genomes and 15-32 more middle-quality genomes than the second-best result. Binning refiner with CLMB and VAMB\cite{nja21} achieves the best performance than any other binners. Impressively, in addition to being compatible with the binning refiner, single CLMB even recovers on average 15 more HQ genomes than the refiner of VAMB and Maxbin on the benchmarking datasets. Furthermore, CLMB is applied to a real mother-infant microbiome dataset with 110 samples and recovers 365 high-quality and middle-quality genomes, including 21 new ones. As a crucial step for metagenomic research, the genome recovered by CLMB provides insights into the microbiome transmission.

Our contributions in this paper are summarized as follows:
\begin{itemize}
    \item We propose a new metagenomic binner, CLMB, based on deep contrastive learning. It is the first binner that can effectively handle the noise in the metagenomic data. By implicitly modeling the noise using contrastive learning, our method can learn stable and robust representations for the contigs, thus leading to better binning results.
    \item We propose a novel data augmentation approach for metagenomic binning under the contrastive learning framework. Experiments suggest that it can indeed help us model the noise implicitly.
    \item We carefully evaluate the contribution of different properties and features to metagenomic binning using our method, including the sequence encoding, dimension, abundance, \textit{etc.} We also show how our method can be combined with other binners to further improve the binning step. It can guide the users to achieve a better binning result. 

\end{itemize}


\section{Methods}
The key idea of CLMB is to involve explicitly modeled noise it in the data, to learn effective contig representations, and to pull together the representations of functionally similar contigs, while pushing apart dissimilar contigs. We achieve the goal with deep contrastive learning. 

\begin{figure}[htpb]
  \begin{center}
    \includegraphics[width=0.95\columnwidth]{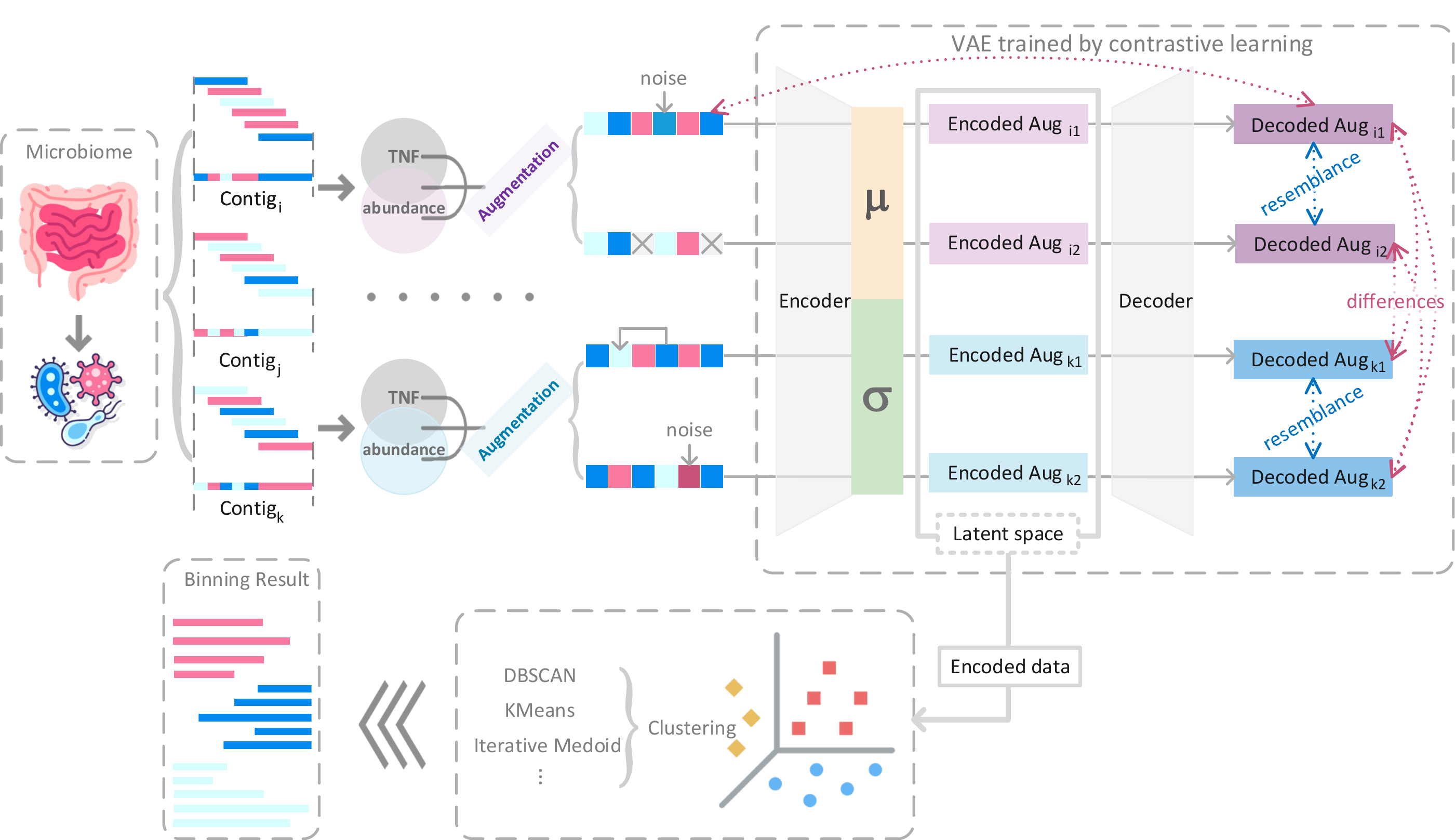}
  \end{center}
\caption{\textbf{Overview of CLMB workflow.} CLMB takes contigs from sampled microbiome as inputs. Then, the abundances and the per-sequence tetranucleotide frequencies (TNF) are calculated, concatenated, and subsequently augmented to a pair of distorted data. All the augmented data are passed through VAE to train it with contrastive learning. After training, the concatenated features of each contig are passed through VAE to obtain the encoded data in the latent space as the representation. Finally, a general clustering algorithm can be applied to the representations to obtain binning results.}
\label{overview}
\end{figure}

The CLMB pipeline is shown in Figure~\ref{overview}.
The inputs of CLMB are the contigs assembled from sequencing reads.
For each contig, the abundances and the per-sequence tetranucleotide frequencies (TNF) are respectively calculated and transformed to numerical vectors of $s$-dimensional and $103$-dimensional, denoted $A_{in}$ and $T_{in}$ (Methods~\ref{m:f} in Appendix, $s$ denotes the number of samples), both of which were concatenated as the input feature, denoted $concat(A_{in},T_{in})$.
Given the feature, we simulate noise in different forms, such as Gaussian noise and random mask, and add the noise to it, resulting in slightly distorted feature as the augmented data.
Specifically, for each contig, two random augmented data are generated based on the feature data (Section \ref{2d}) and used to train a neural network with contrastive learning, \textit{i.e.}, contrasting the training pair of each contig between each other and against other data pairs \cite{hk20}.
As for the neural network model, we select the variational autoencoder (Section \ref{2a}), due to its capability of learning smooth latent state representations of the input data \cite{kw14,rmw14}.
When training the VAE model (Section \ref{2t}), we force the model to produce similar representations for the augmented data of the same contig while distinct for those of different contigs (contrastive learning).
More specifically, by discriminating the augmented data of the same contig from massive augmented data of the other contigs, the deep neural network (VAE) parameterizes a locally smooth nonlinear function $f_{\theta}$ that pulls together multiple distortions of a contig in the latent space and pushes away those of the other contigs.
Intuitively, as the representations of the augmented data from the same contig are pulled together by $f_{\theta}$, contigs with similar feature data can be pulled together in the latent space, which are more likely to be placed in the same cluster. 
After contrastive learning, $concat(A_{in},T_{in})$ of each contig can be encoded by the trained VAE to the mean of their denoised distributions in the latent space (Section \ref{2e}).
The mean data of the contigs are the representations that we learn, which are
subsequently clustered with the common clustering algorithms (\textit{e.g.}, minibatch k-means \cite{scu10}, DBSCAN \cite{eksx96}, iterative medoid clustering algorithm \cite{kfem15,nja21})\footnote{Minibatch k-means and DBSCAN are implemented by scikit-learn: https://scikit-learn.org . Iterative medoid clustering algorithm are implemented by \cite{nja21}: https://github.com/RasmussenLab/vamb/blob/master/doc/tutorial.ipynb .
} and put into respective bins (Section \ref{2e}).

\subsection{Data augmentation}
\label{2d}

Data augmentation is essentially the process of modeling the noise explicitly.
Any noise in real-world conditions that influence the quality of metagenomic short reads might result in the implicit change of feature data.
For example, base deletion during genomic sequencing causes a statistical error of the tetramer frequencies and consequently the distortion of TNFs.
Therefore, we perform data augmentation to the feature data for interpretability and effectiveness.
We design three augmentation approaches for three noise cases, considering the real-life metagenoimc sequencing and data analytic pipeline.
\begin{enumerate}
    \item Gaussian noise.
    It simulates the unexpected noise in metagenomic sequences.
    Assuming the features conform to Gaussian distribution with mean $\mu$ and variance $\sigma^2$, the noise obtained by sampling the Gaussian distribution $N(0,\sigma^2)$ and scaled in $0.15\mu$ is added to the feature data.
    \item Random mask.
    This simulates undetected read mapping during the assembly.
    Each dimension of the feature data might be masked with 0.01 probability.
    \item Random shift.
    This kind of noise covers the imperfect genomic sequencing process.
    Two dimensions, $i$ and$j$, of the feature data are chosen, and the number $f[i]$ on dimension i turns into $\frac{9f[i]}{10}$ while the number $f[j]$ on dimension j turns into $f[j]+\frac{f[i]}{10}$.
    The total percentage of chosen pairs of dimension is 0.01.
\end{enumerate}

Three approaches make up 6 augmented form pairs in total, and one of them is randomly selected for each data augmentation during training, generating training pairs for the feature data of each contig.
After this, a minibatch of $N$ contigs generates the augmented data with size $2*N$.

\subsection{Architecture of the VAE}
\label{2a}

We employ the VAE architecture constructed in \cite{nja21}.
For a minibatch of $N$ contigs, augmented data with size $2*N$ are passed through the VAE module.
Each $(s+103)$-dimensional vector, generated from the augmentation of $concat(A_{in},T_{in})$, is firstly passed through two fully connected layers with batch normalization \cite{is15} and dropout (P=0.2) \cite{hskss12}, termed the encoding layers, parameterizing function $f_e$.
The output of the last layer, with $N_h$ dimension, is then passed to two different fully connected layers with $N_h$ dimensions, termed the $\mu$ and $\sigma$ layers, parameterizing function $f_{\mu}$ and $f_{\sigma}$, respectively.
The latent layer, $l$, is obtained by sampling the Gaussian distribution using the $\mu$ and $\sigma$ layers as parameters, \textit{i.e.}, $l_i\sim N(\mu_i, \sigma_i)$ for each neuron $i=1,2,...,N_h$.
The sampled latent representation is then passed through the decoding layers, with the same size as the encoding layers except for arranged in a reverse order, parameterizing function $f_d$.
Followed by the last decoding layer is a fully connected layer of $s+103$ dimensions with function $f_s$ parameterized, in which the vector is splited into two output vectors of dimension $s$ and 103, $A_{out}$ and $T_{out}$, as the output abundance and TNFs, respectively.
We use linear activation for the $\mu$ layer, softplus activation for the $\sigma$ layer, and leaky rectified linear activation \cite{mmhn13} for the other layers. 

\subsection{Loss function}
\label{2l}
The loss function of CLMB is a trade-off for three goals: 
\begin{enumerate}
    \item The decoded data should be similar to the input data, which is a  requirement of training autoencoder;
    \item The Gaussian distribution dependent on the $\mu$ and $\sigma$ layers for sampling is constrained by a prior $N(0,I)$, which is the prerequisite of VAE \cite{kw14,rmw14}.
    \item The decoded data for the augmented data of the same contig are as similar as possible, while those of different contigs are as dissimilar as possible, which is the terminal condition of contrastve learning \cite{hk20}.
\end{enumerate}

To satisfy the first goal, we have
\begin{equation}
\label{loss1}
    L_1=w_A \sum ln(A_{out}+10^{-9}) \cdot A_{in}+w_T \sum (T_{out}-T_{in})^2,
\end{equation}
where the $w_A$ and $w_T$ are the weighting terms. We use cross-entropy to penalize the abundance bias and the sum of squared errors to penalize the TNFs bias. 

To satisfy the second goal, we have
\begin{equation}
\label{loss2}
    L_2=- \sum \frac{1}{2} (1+ln(\sigma)-\mu^2-\sigma)~~~~~~ \cite{doersch2021tutorial,hk20}.
\end{equation}
We use the Kullback–Leibler divergence to penalize the deviance from this distribution.

To satisfy the third goal, we investigate the structure of each minibatch of $2*N$ (distorted) augmented data, which are obtained by performing data augmentation to $\{concat(A_{in},T_{in})_k\}^{N}_{k=1}$ of $N$ contigs. All the data are passed through the VAE module, and we denote the output data from the decoding layer as $X=\{x_k \in R^{s+103}\}^{2N}_{k=1}$.
For a pair of positive data $x_i$ and $x_j$ (derived from the feature data of the same contig), the other $2*N-2$ samples are treated as negatives.
To distinguish the positive pair from the negatives, we define the cosine distance between two vectors $cos(x_i,x_j)=\frac{x_i^T \cdot x_j}{||x_i||\cdot ||x_j||}$ and use the normalized temperature-scaled cross-entropy loss:
\begin{equation}
\label{lossn}
    l_{i,j}=-log\frac{e^{\frac{cos(x_i,x_j)}{\tau}}}{\sum^{2N}_{s=1,s\neq i} e^{\frac{cos(x_i,x_s)}{\tau}}},
\end{equation}
where the temperature $\tau$ is a parameter we can tune.
Note that $l(i,j)$ is asymmetrical.
Suppose all the pairs $X=\{x_k \in R^{s+103}\}^{2N}_{k=1}$ are put in an order, in which $x_{2k-1}$ and $x_{2k}$ denote a pair of positive data, the summed-up loss within this minibatch is:
\begin{equation}
\label{loss3}
    L_3=\frac{1}{2N} \sum^{N}_{k=1} (l_{2k-1,2k}+l_{2k,2k-1}).
\end{equation}

Finally, the combined loss function is
\begin{equation}
\label{loss}
    LOSS=L_1+w_2L_2+w_3L_3.
\end{equation}
The weighting terms are set as $w_A=0.85 ln(s)^{-1}$, $w_T=0.15/103$, $\tau=0.1$, $w_2=\frac{L_{1(0)}/L_{2(0)}}{2\times 10^5 N_h}$, $w_3=1.35 L_{1(0)}/L_{3(0)}$, where $L_{1(0)} ,L_{2(0)},L_{3(0)}$ indicate the value of $L_1,L_2,L_3$ at the first epoch and are initially set to 1.

\subsection{Training with contrastive learning}
\label{2t}
Here, we have modelled the noise explicitly, constructed the architecture, and defined the loss function we should optimize.
The contrastive learning algorithm for training process will force the architecture to be robust to the noise we modelled. The pseudocode for training is presented in Algorithm~\ref{alg}.

\begin{algorithm}[ht]
    \caption{The contrastive learning algorithm for training VAE}
    \label{alg}
	\hspace*{0.02in}{\bf Input:}
	\quad batchsize $N$, constant parameter $\tau$, structure of $f_e,f_{\mu},f_{\sigma},f_d,f_s$, feature data $concat(A_{in},T_{in})$
	
	
	\begin{algorithmic}[1]
        \FOR{sampled minibatch $\{concat(A_{in},T_{in})_k\}^{N}_{k=1}$}
            \STATE{select one data augmentation form pair with augmentation functions $t_1,t_2$;}
            \FOR{all $k\in\{1,2,...,k\}$}
                \STATE{$Aug_{2k-1}=t_1(concat(A_{in},T_{in})_k)$;~~$Aug_{2k}=t_2(concat(A_{in},T_{in})_k)$~~~~~~~~\textit{\#Augmentation}}
                \STATE{$\mu_{2k-1}=f_{\mu}(f_e(Aug_{2k-1}));~~\mu_{2k}=f_{\mu}(f_e(Aug_{2k}))$}
                \STATE{$\sigma_{2k-1}=f_{\sigma}(f_e(Aug_{2k-1}));~~\sigma_{2k}=f_{\sigma}(f_e(Aug_{2k}))$}
                \STATE{sample $l_{2k-1}$, $l_{2k}$ from the multivariate gaussian distribution $N(\mu_{2k-1}, \sigma_{2k-1})$, $N(\mu_{2k}, \sigma_{2k})$ respectively.~~~~~~~~~~~~~~~~~~~~~~~~~~~~~~~~~~~~~~~~~~~~~~~~~~~~~~~~~~~~~~~~~~~~~~~~\textit{\#Representation}}
                \STATE{$x_{2k-1}=f_d(l_{2k-1});~~x_{2k}=f_d(l_{2k})$~~~~~~~~~~~~~~~~~~~~~~~~~~~~~~~~~~~~~~~~~~~~~~~~~~~~~~~~\textit{\#Projection}}
                \STATE{$A_{out_{2k-1}},T_{out_{2k-1}}=f_s(x_{2k-1});~~A_{out_{2k}},T_{out_{2k}}=f_s(x_{2k})$~~~~~~~~~~~~~~~~~~~~~~~~~~~~~\textit{\#Splitting}}
            \ENDFOR
            \STATE{$L_1=w_A \sum_{k=1}^{2N} ln(A_{out_k}+10^{-9}) \cdot A_{in_{\lfloor \frac{k+1}{2} \rfloor}}+w_T \sum (T_{out}-T_{in_{\lfloor \frac{k+1}{2} \rfloor}})^2$}
            \STATE{$L_2=- \sum_{k=1}^{2N} \frac{1}{2} (1+ln(\sigma_k)-\mu^2_k-\sigma_k)$}
            \STATE{$L_3=\frac{1}{2N} \sum^{N}_{k=1} (l_{2k-1,2k}+l_{2k,2k-1})$, where $l_{i,j}$ is defined in Equation~\ref{lossn}}
            \IF{in the first epoch and $w_2=w_3=1$}
                \STATE{calculate $w_2,w_3$ based on the value of $L_1,L_2,L_3$}
            \ENDIF
            \STATE{$LOSS=L_1+w_2L_2+w_3L_3$}
            \STATE{update networks $f_e,f_{\mu},f_{\sigma},f_d,f_s$ to minimize LOSS}
        \ENDFOR
		\RETURN{encoding structure $f_e,f_{\mu}$}
	\end{algorithmic}
\end{algorithm}

As shown in Algorithm~\ref{alg}, in each training epoch, the contigs are randomly seperated to several minibatches. The augmented data of each minibatch are put into VAE for training. The loss function is determined after  $L_1,L_2,L_3$ are calculated. We train VAE by optimizing $LOSS$ using the Adam optimizer \cite{kb17} and using one Monte Carlo sample of the Gaussian latent representation.

Algorithm~\ref{alg} trains VAE by discriminating the data in sampled minibatch. However, due to insufficient memory capacity (either of CPU or GPU), a limited proportion of data are sampled to a minibatch, which might lead to a problem that the VAE fits well with the data in the minibatch rather than the whole dataset.
Therefore, contrastive learning can benefit from shuffled, larger batch size and more epoches for training \cite{hk20}.
We train the model with minibatches of 4096 contigs for 600 epoches.

\subsection{Productive model}
\label{2e}
After training, we define the productive function $f_\theta(x)=f_{\mu}(f_e(x))$, \textit{i.e}, the mapping parameterized by the encoder layers connected with the $\mu$ layers.
Therefore, given the feature data $concat(A_{in},T_{in})$ of a contig, we obtain the representations $f_{\mu}(f_e(concat(A_{in},T_{in})))$ by passing the data through the the encoder layers and the $\mu$ layers.
Once we obtain the representations of all the contigs, we cluster them with the common clustering algorithms (\textit{e.g.}, minibatch k-means \cite{scu10}, DBSCAN \cite{eksx96}).
We find that the iterative medoid clustering algorithm developed by \cite{nja21} is the state-of-art clustering algorithm specifically for metagenome binning (Figure~\ref{dc} in Appendix). After clustering, contigs in the same cluster are put into the same bin. Moreover, for the multisplit workflow, the contigs in the same bin should also be separated based on their source samples\cite{nja21}.


\section{Results}

\subsection{Datasets and Evaluation metrics.}
\label{da}

~~~~\textbf{Datasets.} 
To show the performance of CLMB, we use the benchmarking datasets, which are five synthetic datasets from the CAMI2 Toy Human Microbiome Project Dataset \cite{shb17}: Airways (10 samples), Gastrointestinal (GI, 10 samples), Oral (10 samples), Skin (10 samples), and Urogenital (Urog, 9 samples)\footnote{You can get the whole package data from https://data.cami-challenge.org/participate, or get the contigs and calculated abundance from https://codeocean.com/capsule/1017583/tree/v1}.
For each dataset, contigs$<2,000$ base pairs are discarded.
We obtain the abundance data in numpy\footnote{https://numpy.org} format from the website of \cite{nja21}\footnote{https://codeocean.com/capsule/1017583/tree/v1}, which are calculated using $jgi\_summarize\_bam\_contig\_depths$, implemented by \cite{klkteaw19} on BAM files created with bwa-mem \cite{ld09} and sorted with samtools \cite{l09}.

\textbf{Evaluation metrics.}
We adopt the evaluation metrics for taxonomic binning defined in \cite{shb17} as done in previous work \cite{klkteaw19,nja21}.
After the bins are obtained, we match each bin with each reference genome.
We define the number of nucleotides in the genome covered by contigs from the bin as true positives (TP); 
the number of nucleotides from other genomes covered by contigs in the bin as the false positives (FP); 
the number of nucleotides in the genome covered by contigs in the dataset, but not by any contig in the bin as the false negatives (FN).
Then, $Precision=\frac{TP}{TP+FP}$ and $Recall=\frac{TP}{TP+FN}$ are calculated.
All the CAMI2 datasets have taxonomy files with the definition of strain, species, and genus taxonomic levels.

\subsection{CLMB recovers more near-complete genomes on most datasets}
\label{nc}

\begin{figure}[htpb]
  \begin{center}
    \includegraphics[width=0.95\columnwidth]{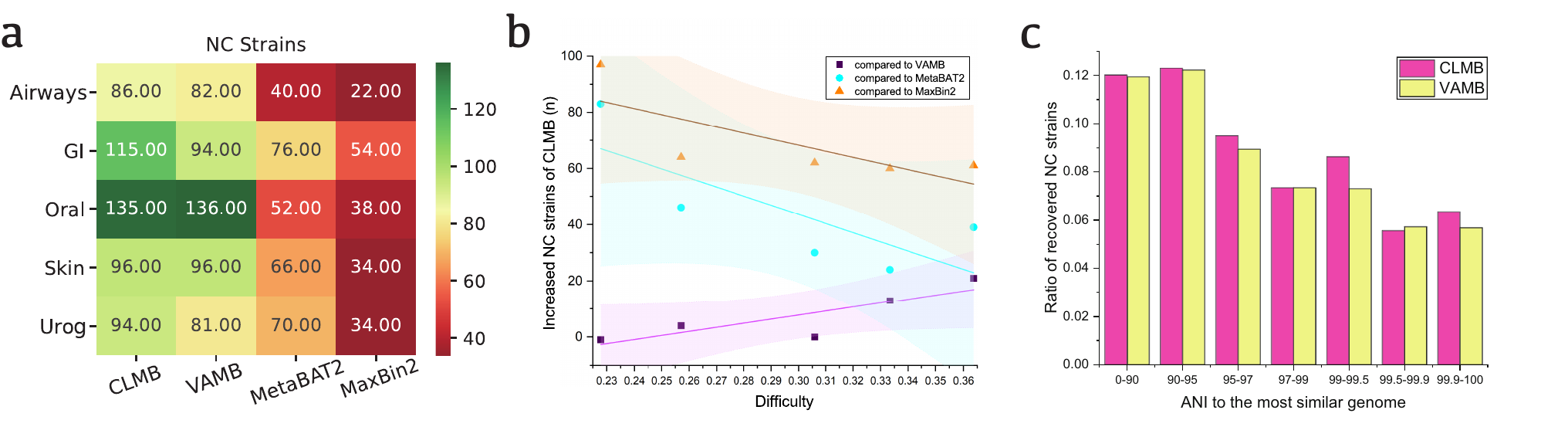}
  \end{center}
\caption{\textbf{Performance comparison on benchmarking datasets.} \textbf{a}. Number of NC strains recovered from the six benchmarking datasets for CLMB, VAMB, MetaBAT2, and MaxBin2. \textbf{b}. The linear fitting and 95\% confidence interval of the difficulty of the dataset and the increased number of NC strains recovered by CLMB relative to VAMB (Pearson correlation coefficient=0.85), MetaBAT2 (Pearson correlation coefficient=-0.77) and MaxBin2 (Pearson correlation coefficient=-0.75). The difficulty is defined as the reciprocal of the Shannon entropy (see Supplementary Table 1 from \cite{nja21}) of the dataset and is always positive. \textbf{c}. The ratio of recovered NC genomes to total reference genomes (which is regarded as ideally recoverable genomes), divided by the ANI to the most similar reference genomes across CAMI2 datasets. CLMB, pink; VAMB, yellow.}
\label{1}
\end{figure}

We ran CLMB on the five CAMI2 datasets.
For each dataset, the augmented data serve as training data, while the original feature data serve as testing data.
Therefore, CLMB obtains a specific encoding function $f_\theta$ parameterized by VAE for each dataset.
In addition, as the data augmentation is performed several times during training, CLMB has a larger data volume for training than the input data volume.

We also benchmarked VAMB \cite{nja21}, MetaBAT2 \cite{kfem15} and Maxbin2 \cite{wss16} on the five benchmarking datasets for comparison.
We evaluated the binning performance by the number of recovered Near-Complete (NC, $recall>90\%$ and $precision>95\%$) genomes as the previous works \cite{nja21,shb17,b17}.
Firstly, CLMB reconstructed 4-21 more NC genomes at the strain level over the second-best binners on three of the five benchmarking datasets (Airways, GI, Urog), and equivalent NC strains to VAMB on Skin and Oral datasets (Figure~\ref{1}a and Table~\ref{tab:st} in Appendix).
Secondly, the increased performance of CLMB relative to MetaBAT2 and Maxbin2 is very significant. Moreover, the increased performance of CLMB to VAMB is positively correlated with the difficulty of the CAMI2 datasets (which is defined as the reciprocal of the Shannon entropy of the datasets\footnote{The Shannon entropy of the five datasets are calculated by \cite{nja21} on their Supplementary Table 1.} because higher Shannon entropy indicates more information contained in the dataset and lower difficulty for binning.) (Figure~\ref{1}b). That indicates that our method indeed resolves the bottleneck of the other methods when the dataset becomes more noisy and difficult. More specifically, CLMB reconstructed more NC strains for most datasets compared to MetaBAT2 and Maxbin2.
Compared to VAMB, CLMB reconstructed more NC strains for high-difficulty datasets and approximately equivalent NC strains for low-difficulty datasets.
Thirdly, CLMB reconstructed on average 10\% more species under any criteria for the GI and Urog datasets, and 8\% more species under stricter criteria (e.g., $Recall>0.90$) for the Airways and Skin datasets.
However, if loosening the criterion (e.g., $Recall>0.70$), CLMB reconstructed 1\%-5\% fewer species on Airways and Skin datasets than VAMB, which had similar performance to CLMB on the Oral dataset with VAMB 0.5\% better across all the criteria except for $Recall>0.99$ (Table~\ref{tab:s} in Appendix).
At the genus level, CLMB outperformed VAMB on datasets Airways, GI, Oral, Skin under stricter criteria, but on the contrary under looser criteria.
On the Urog dataset, CLMB was the second-best binner, recovering approximately 10\% fewer genus than MetaBAT2 (Table~\ref{tab:g} in Appendix), which is a meta-binner.

We further mapped the recovered genomes to reference genomes and counted the average nucleotide identity (ANI) between each reference genome.
Ideally, all the reference genomes are recovered after the sequencing, assembly, and binning process, which is, however, extremely hard in real-world conditions.
For each reference genome, we found the most similar genome and counted the ANI between them.
The NC genomes recovered by CLMB can be mapped to 6\% of all reference genomes having $>99.9\%$ ANI to the most similar genome (Figure~\ref{1}c).
Moreover, compared to VAMB, the NC genomes recovered by CLMB were mapped to more reference genomes across all the intervals of ANI except for 99.5\%-99.9\% ANI.

\subsection{The performance of CLMB benefits from finding the information of resemblance and discrimination within data}
\label{fe}

\begin{figure}[htp]
  \begin{center}
    \includegraphics[width=0.95\columnwidth]{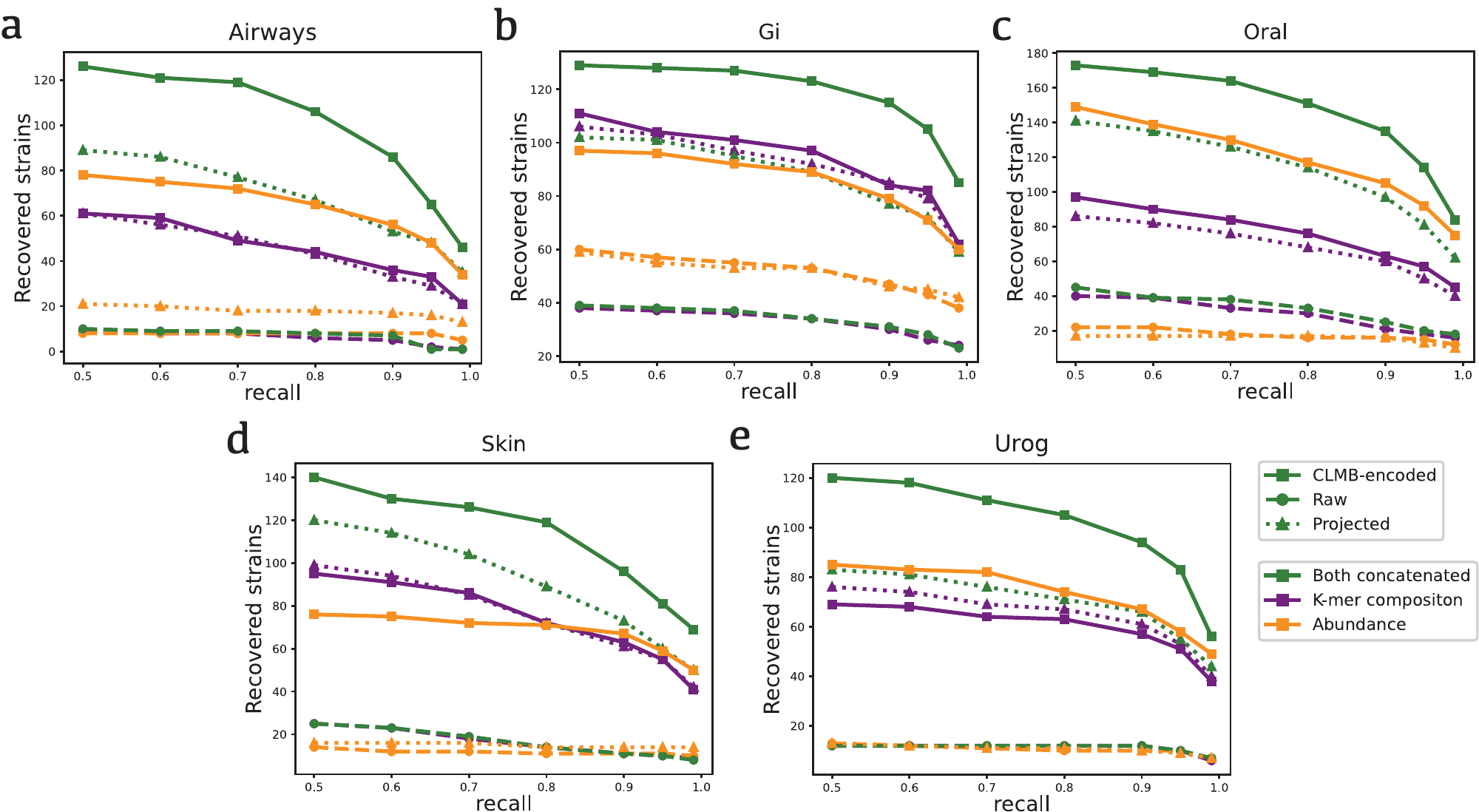}
  \end{center}
\caption{\textbf{Results of data fusion experiments}. Fusion test of 5 benchmarking datasets for CLMB, precision=0.95, recall range from 0.5 to 0.99.
Color: Abundance (Yellow), k-mer composition (Purple),  both concatenated(Green)
Linestyle: Raw data (Round), Projected data (triangle), CLMB-encoded data (square)}.
\label{2}
\end{figure}

~~~~We conducted the data fusion experiment \cite{haa16} on the five CAMI2 challenging datasets, \textit{i.e}, comparing the performance of the abundance, k-mer composition, or both concatenated.
Because the representation of all the data encoded by CLMB would be projected to 32-dimension space by $f_e,f_{\mu}$ (Figure~\ref{overview}), we also projected raw data to 32-dimension space using Principal Components Analysis (PCA) \cite{c16}, termed as ‘projected data’, to avoid the clustering results affected by different dimensions.
We tested the number of NC strains produced by binning with raw data, projected data, and CLMB-encoded data in the data fusion experiment, respectively (Figure~\ref{2}).

On datasets Airways, Oral, Skin, and Urog, the raw data of both concatenated did not achieve better results than the raw data of single abundance or single k-mer composition, but the projected data of both concatenated yielded 5\%-700\% more genomes than that of single data.
This interesting result proved that the dimension of input data did affect the clustering and binning result, and more information contained in the concatenated data was beneficial to the clustering result after eliminating the variation of dimensionality.
On dataset GI, the raw data of both concatenated achieved worse results than the raw data of single abundance, but the projected data of both concatenated yielded worse results than the single k-mer composition.
This might stem from the information conflict between k-mer composition and abundance.
With contrastive learning, the three CLMB-encoded data recovered 3-12 times more NC genomes than the corresponding raw data.
Moreover, the CLMB-encoded data of both concatenated and abundance also recovered on average 19\% and 189\% more than the projected data ones, although the CLMB-encoded data of k-mer composition had similar performance to the projected data.
Most importantly, the CLMB-encoded data of both concatenated achieved the best performance across all the datasets, recovering on average 17\% more genomes than the second-best results, no matter what performance the raw data or projected data of both concatenated achieved.

\begin{figure}[htpb]
  \begin{center}
    \includegraphics[width=0.9\columnwidth]{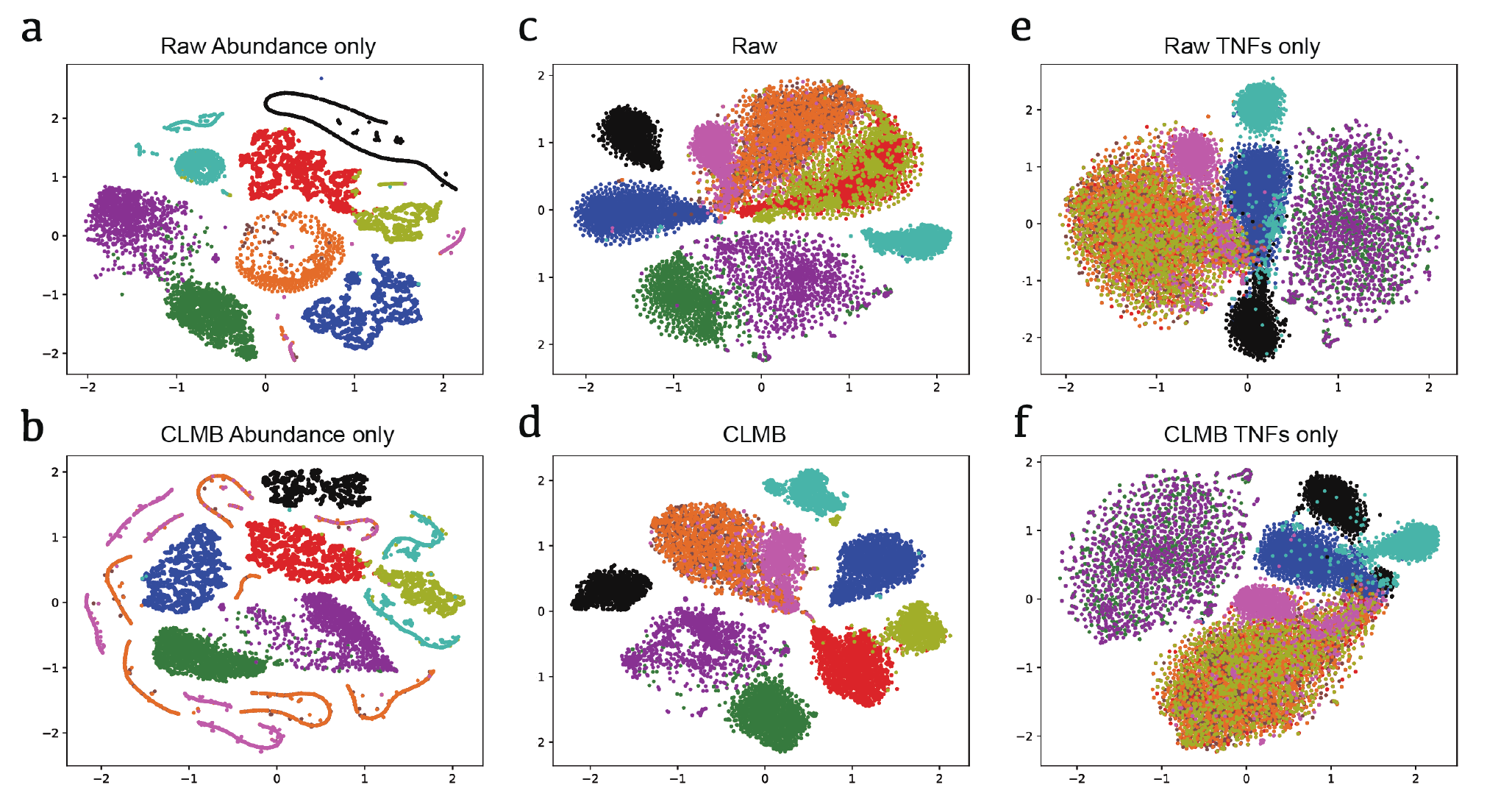}
  \end{center}
\caption{\textbf{T-SNE visualiztion of data fusion experiments on Skin dataset.}
We randomly selected 10 of 15 strains with maximum contigs from the CAMI2 Skin dataset. Each point represents a contig from that strain, and points with same color means originating from same strain (\textit{i.e}, the same reference genome). \textbf{a,b}. Raw data (a) and CLMB-encoded data (b) of abundance. \textbf{c,d}. Raw data (c) and CLMB-encoded data (d) of both concatenated. \textbf{e,f}. Raw data (e) and CLMB-encoded data (f) of k-mer composition.}
\label{v}
\end{figure}

We also visualized the raw data and CLMB-encoded data on dataset Skin, using t-SNE\cite{mh08} (Figure~\ref{v}).
Firstly, the CLMB-encoded data of both concatenated appeared to have genomes more clearly separated than any other cases.
Figure~\ref{v}a, \ref{v}c, and \ref{v}e showed that, more information contained in both concatenated contributed little to the cluster separation, which is similar to the result of the data fusion experiment.
However, Figure~\ref{v}b, \ref{v}d, and \ref{v}f showed that, the CLMB-encoded data of both concatenated appeared to have genomes more clearly separated than any other cases.
It suggests that CLMB leverages the information within data to achieve better performance.

Furthermore, the performance of CLMB-encoded data of both concatenated was dependent on the number of selected samples (which decided the dimension of the abundance) (Figure~\ref{ds} in Appendix). Another experiment tested the effect of different k (2–5) for encoding k-mers composition, and in accordance with empirical results \cite{kbd09,pmwb03}, showed that k=4 gave the best or second performance on all the datasets (Figure~\ref{dk} in Appendix).

\subsection{The performance of the ensemble binning is improved by involving CLMB}
\label{eb}

\begin{figure}[htbp]
  \begin{center}
    \includegraphics[width=0.9\columnwidth]{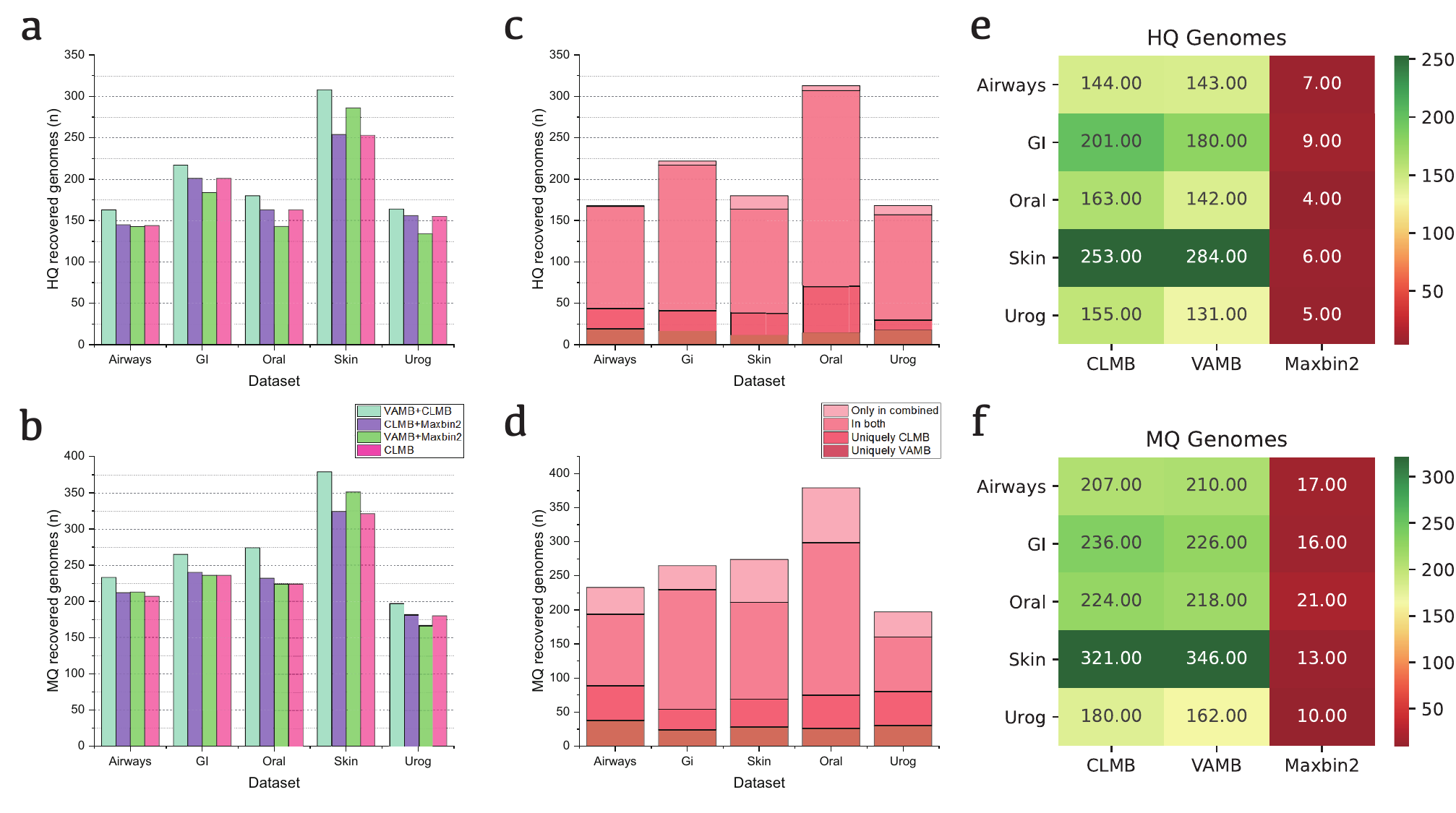}
  \end{center}
\caption{\textbf{Quality assessment of genomes recovered by binners.} \textbf{a, b}. The number of high-quality (a) and middle-quality (b) genomes obtained using MetaWRAP binning refinement tool. We used the binning result from 1) CLMB and VAMB (light cyan), 2) CLMB and MaxBin2 (purple), and 3) VAMB and MaxBin2 (green). The number of high-quality (a) and middle-quality (b) genomes recovered by a single CLMB (pink) is used for comparison.
\textbf{c, d}. The source of the results of the MetaWRAP binning refinement tool. We investigated the number of HQ (c) and MQ (d) genomes uniquely from one of the two binners (dark pink, medium pink), found in both binners (light pink), and the number of genomes that were not HQ (c) or MQ (d) in any binner but were regenerated as HQ (c) or MQ (d) in the binning refinement output (lightest pink).
\textbf{e, f}. Number of HQ (e) and MQ (f) genomes recovered by single CLMB, single VAMB, and single MaxBin2.}
\label{3}
\end{figure}

The ensemble binning refinement method is popular after draft metagenome binning because they combine bins from multiple programs. To show that CLMB is compatible with the ensemble binning tool, we ran MetaWRAP bin-refinement \cite{udt18,sb73-5} on the five CAMI2 challenging datasets by involving CLMB.
Because MetaWRAP bin refiner used CheckM \cite{pisht15} to assess the quality of recovered genomes, we here evaluated the performance by the number of recovered high-quality (HQ, $completeness>90\%$ and $contamination<5\%$) genomes or middle-quality (MQ, $50\%<completeness<90\%$ and $contamination<5\%$) genomes as the previous works \cite{f18,p19}. 
The bin refiner of two binners usually outperformed single binner, and the refiner of CLMB and VAMB performed best, recovering 8-22 more  HQ genomes and 15-32 more MQ genomes than the second-best method.
We also found that the refiner of CLMB and Maxbin2 outperformed that of VAMB and Maxbin2 on four of five datasets (Figure~\ref{3}~a,b). Moreover, CLMB and VAMB agreed on over a half of the HQ genomes and MQ genomes, but CLMB recovered more unique HQ genomes on average (Figure~\ref{3}~c,d).

Notice that the CheckM results are not equivalent to the benchmarking results for each binner, which is due to different evaluation methods. We then revisited the benchmarking experiments except for evaluating the performance by the number of recovered HQ genomes and MQ genomes.
On datasets GI, Oral, and Urog, CLMB recovered 21-22 more HQ genomes or 6-18 more MQ genomes than VAMB, which had similar performance to CLMB on Airways and better performance than CLMB on Skin (Figure~\ref{3}~e,f).
Impressively, on datasets Airways, GI, Oral, and Urog, single CLMB even recovered on average 15 more HQ genomes than the refiner of VAMB and Maxbin (Figure~\ref{3}~a,b).

In conclusion, the performance of binning refiner is highly dependent on the performance of all the involved binners.
As many metagenomics studies screen the bins based on their quality after metagenome binning for future analysis, we expect that more HQ and MQ genomes can be distinguished using CLMB and the binning refinement methods.

\subsection{The genomes recovered by CLMB assist analysis for mother-infant microbiome}
\label{mi}

\begin{figure}[htbp]
  \begin{center}
    \includegraphics[width=0.9\columnwidth]{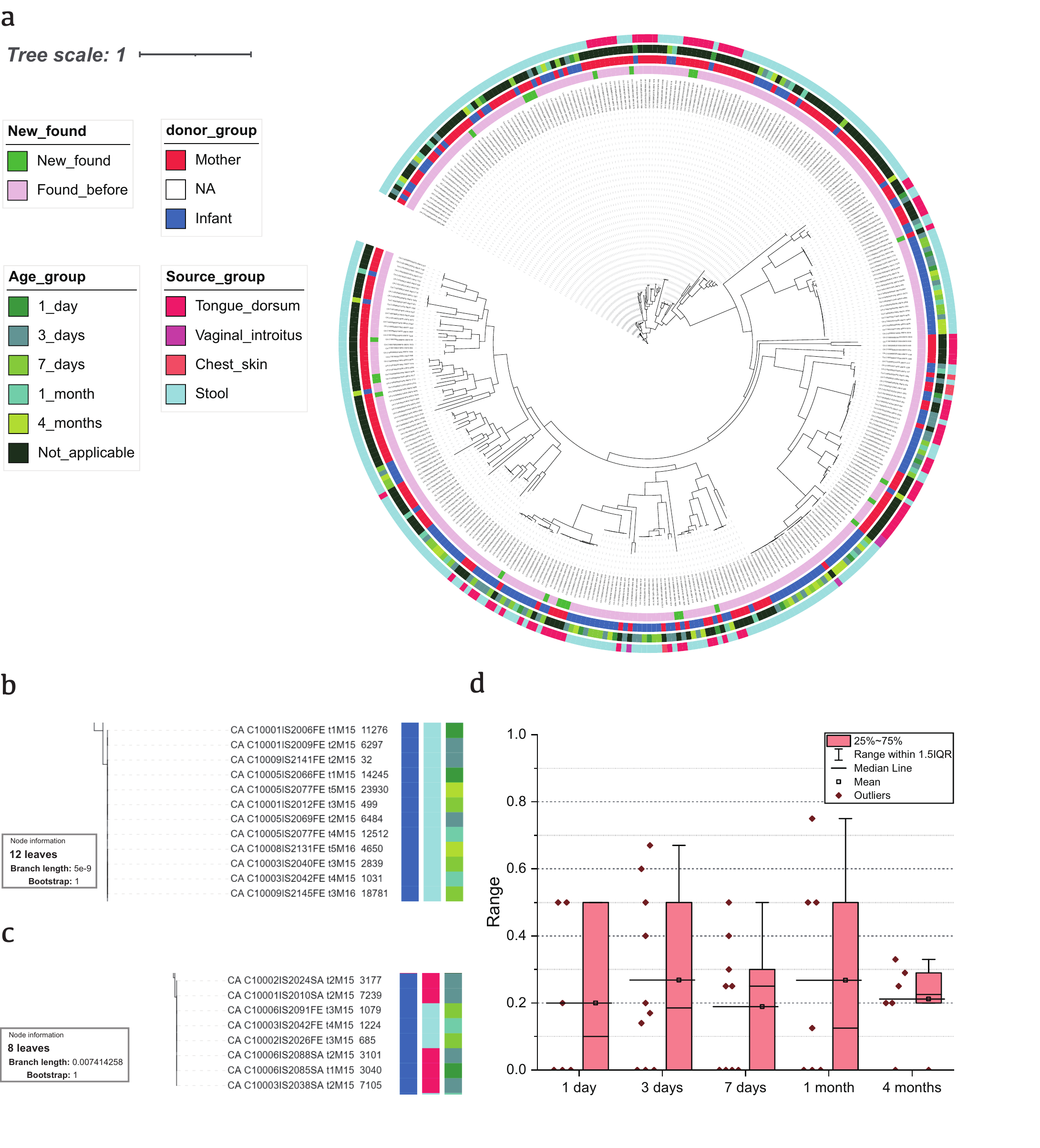}
  \end{center}
\caption{\textbf{Metagenomic analysis on mother-infant microbiome.} \textbf{a}. Cladogram of species tree of all the 365 bins generated. The annotation rings, from inner to outer: 1) the bins of new-found strains (green) or discovered before (light pink) in \cite{f18}; 2) the sample is donated by mother(red) or infant(blue); 3) the age of infant donor, 1 day (onion green), 3 days (dark green), 7 days (olive green), 1 month (cyan) or 4 months (yellow-green). Not applicable (dark) for mother donors; 4) which human body site the sample is collected from, tongue dorsum (pink), vaginal introitus (lighter pink), chest skin (vermeil), or stool (light cyan).
\textbf{b, c}. The metadata annotations of bins classified as (b) strain \textit{Escherichia coli} and (c) strain \textit{Rothia sp902373285}.
\textbf{d}. The ratio of exclusive species to the total number of species in infants' microbiome. The samples, which obtain 0 species, are not considered.}
\label{4}
\end{figure}

~~~~\textbf{Experiment datasets.} Unlike the above experiments on synthetic datasets, we apply CLMB to real-world data to test the scalability and practicability in this section.
We use the longitudinally sampled microbiome of mother-infant pairs across multiple body sites from birth up to 4 months postpartum from \cite{f18}, which are available at the NCBI Sequence Read Archive (SRA) \cite{lss11} under BioProject number PRJNA352475 and SRA accession number SRP100409.
We select 10 mother-infant pairs with 110 samples and 496342 contigs in total for this experiment.


We ran CLMB on the dataset with default parameters.
We recovered 365 (HQ+MQ) genomes, in which there are 21 new-found strains consisting of 24 bins.
We then reconstructed the phylogeny of all (HQ+MQ) genomes and obtained the unrooted tree \cite{cmhp20}, which are annotated with the metadata file (Figure~\ref{4}~a).
The new-found strains, as annotated, are more from samples of mothers.
We also found that the microbiome of the infants shared more species.
For example, 12 stool samples from 5 infants share strain \textit{Escherichia coli} across ages from 1 day to 4 months, and 8 samples collected from stool and tongue dorsum of 4 infants contain strain \textit{Rothia sp902373285} across ages from 1 day to 1 month (Figure~\ref{4}~b,c).
On the contrary, few strains are shared among mothers in the tree. 
Moreover, the range of strains reconstructed in mothers' samples overlaps little with the range of strains reconstructed in infants' samples.
More than half of the bins are recovered from stool samples, probably because of the larger sequencing files obtained from stool samples than those obtained from samples of other sources (human body sites).
We then counted the newly exclusive strains of the 10 infants.
We found that the proportion of exclusive species has largely changed as they grew up (Figure~\ref{4}~d).
At the age of 4 months, the proportions of exclusive species are within a small range, indicating most infants contained 20\%-30\% exclusive strains found in their microbiome. We suppose that the speed of strain replacement slowed down at that time.


\section{Discussions}
Here, by conciously handling the noise occured in metagenome research, we show improvements on benchmarking datasets.
The improvements, as we have shown, benefit not only from the dimensionality reduction, but also from the model trained by the contrastive learning framework and its robustness to noise.
Furthermore, experiments and applications on real-world datasets demonstrate the scalability and practicability of CLMB.

From the algorithm perspective, CLMB can handle the numerical data that potentially contain error \cite{li2018dlbi}, which is not limited to metagenome binning.
CLMB is promising to handle noise, a significant factor that interferes the data precision.
Therefore, we believe that our findings can inspire not only the field of metagenomics \cite{li2021hmd}, but also other related fields, like structural and functional fields \cite{li2019deep,chen2020rna,li2020modern,wei2021protein}.

\newpage
\setcounter{table}{0}
\setcounter{figure}{0}
\renewcommand{\thetable}{\Alph{section}\arabic{table}}
\renewcommand\thefigure{\Alph{section}\arabic{figure}}

\section{Appendix}
\appendix

\section{Figures}

\begin{figure}[htbp]
  \begin{center}
    \includegraphics[width=0.95\columnwidth]{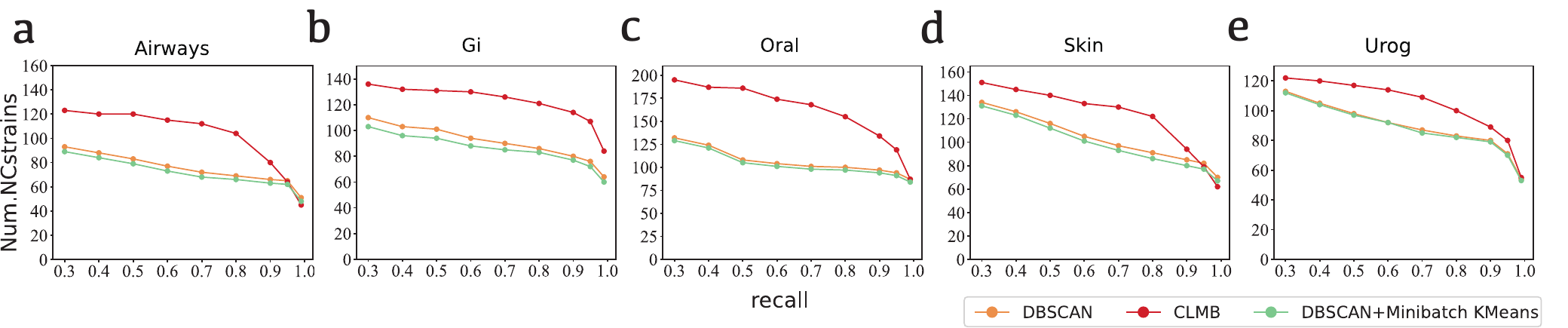}
  \end{center}
\caption{\textbf{Performance of different clustering algorithms based on five datasets.} Orange: DBSCAN Algorithm. Green: Exclude the outlier using DBSCAN first and cluster the others points using minibatch k-means algorithm. Red: Iterative medoid algorithm, which is developed by \cite{nja21} and used by CLMB.}
\label{dc}
\end{figure}

\begin{figure}[htbp]
  \begin{center}
    \includegraphics[width=0.9\columnwidth]{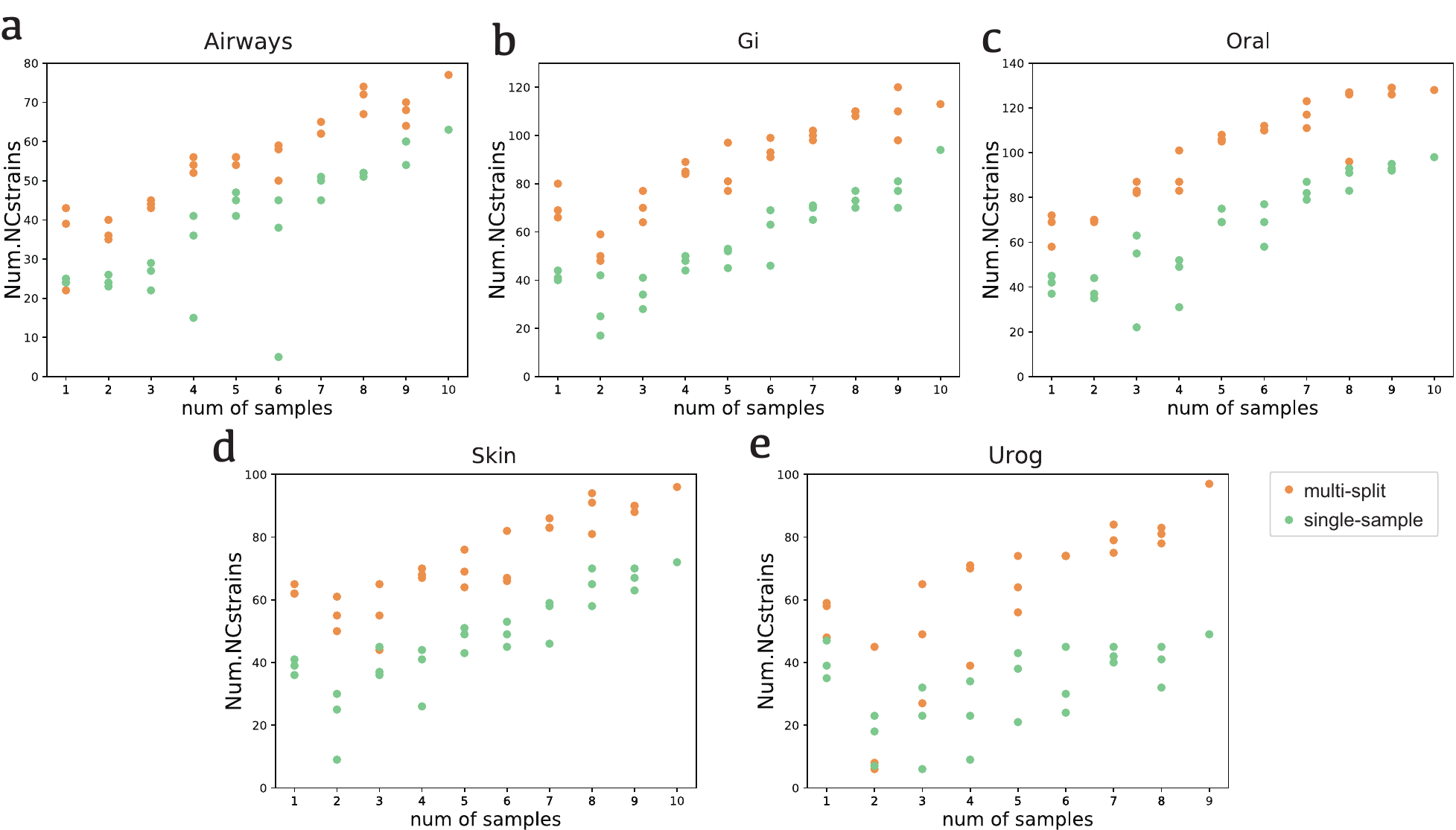}
  \end{center}
\caption{\textbf{Performance of CLMB with different samples.} For any given number of samples, samples were randomly drawn 3 times and executed independently. For “single-sample”, all the samples were run independently. We note that for increasing number of samples, the random subsets chosen is not independent, due to only having 9 (Urog) or 10 (Airways, GI, Skin, Oral) samples in total. Orange: Multi-split workflow of CLMB, Green: Single sample workflow of CLMB.}
\label{ds}
\end{figure}

\begin{figure}[htbp]
  \begin{center}
    \includegraphics[width=0.95\columnwidth]{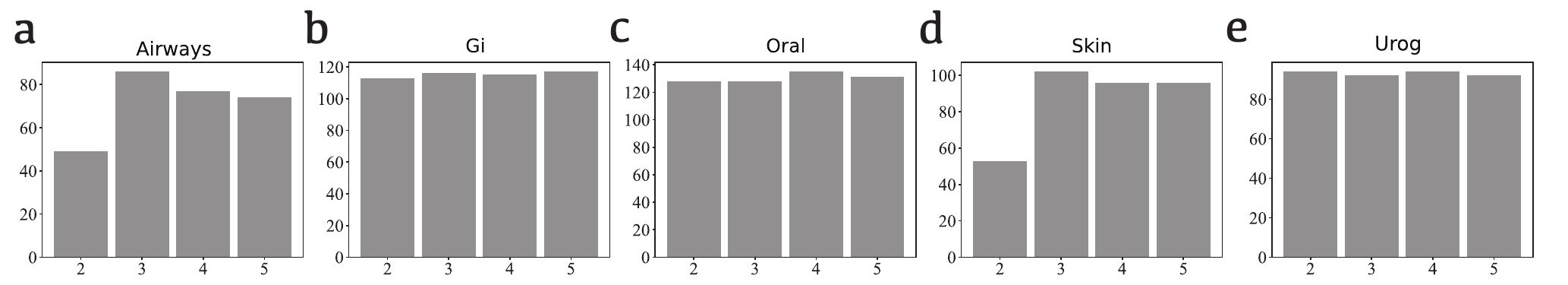}
  \end{center}
\caption{\textbf{Performance of CLMB with different k-mer length on different datasets}. It is assessed by the number of reconstructed NC strains. The performance varies among the datasets.}
\label{dk}
\end{figure}

\newpage

\section{Tables}


\begin{table}[htbp]
  \centering
  \caption{Number of genomes at the strain level reconstructed with a precision of at
least 95\%}
    \begin{tabular}{|c|p{6.68em}|c|c|c|c|c|c|c|}
    \hline
    \multicolumn{2}{|c|}{} & \multicolumn{7}{p{20.065em}|}{RECALL} \bigstrut\\
    \hline
    \multicolumn{1}{|p{6em}|}{Dataset} & Binner & 0.50  & 0.60  & 0.70  & 0.80  & 0.90  & 0.95  & 0.99 \bigstrut\\
    \hline
    \multicolumn{1}{|c|}{\multirow{4}[8]{*}{CAMI2\newline{}Airways}} & MaxBin2 & 42    & 39    & 38    & 33    & 23    & 17    & 13 \bigstrut\\
\cline{2-9}          & MetaBAT2 & 80    & 72    & 66    & 56    & 40    & 30    & 18 \bigstrut\\
\cline{2-9}          & VAMB  & 125   & \textbf{123} & \textbf{120} & \textbf{113} & 79    & 60    & 41 \bigstrut\\
\cline{2-9}          & CLMB  & \textbf{126} & 121   & 119   & 106   & \textbf{86} & \textbf{65} & \textbf{46} \bigstrut\\
    \hline
    \multicolumn{1}{|c|}{\multirow{4}[8]{*}{CAMI2 GI}} & MaxBin2 & 64    & 63    & 63    & 60    & 53    & 50    & 45 \bigstrut\\
\cline{2-9}          & MetaBAT2 & 99    & 97    & 94    & 87    & 76    & 68    & 58 \bigstrut\\
\cline{2-9}          & VAMB  & 121   & 120   & 118   & 113   & 100   & 91    & 77 \bigstrut\\
\cline{2-9}          & CLMB  & \textbf{129} & \textbf{128} & \textbf{127} & \textbf{123} & \textbf{115} & \textbf{105} & \textbf{85} \bigstrut\\
    \hline
    \multicolumn{1}{|c|}{\multirow{4}[8]{*}{CAMI2\newline{}Oral}} & MaxBin2 & 64    & 61    & 55    & 46    & 39    & 31    & 21 \bigstrut\\
\cline{2-9}          & MetaBAT2 & 88    & 86    & 84    & 79    & 73    & 58    & 38 \bigstrut\\
\cline{2-9}          & VAMB  & \textbf{181} & \textbf{174} & \textbf{166} & \textbf{152} & \textbf{135} & 113   & 81 \bigstrut\\
\cline{2-9}          & CLMB  & 173   & 169   & 164   & 151   & \textbf{135} & \textbf{114} & \textbf{84} \bigstrut\\
    \hline
    \multicolumn{1}{|c|}{\multirow{4}[8]{*}{CAMI2\newline{}Skin}} & MaxBin2 & 56    & 53    & 50    & 46    & 34    & 30    & 27 \bigstrut\\
\cline{2-9}          & MetaBAT2 & 106   & 98    & 93    & 76    & 65    & 53    & 42 \bigstrut\\
\cline{2-9}          & VAMB  & 139   & \textbf{133} & \textbf{129} & 116   & \textbf{97} & 80    & 63 \bigstrut\\
\cline{2-9}          & CLMB  & \textbf{140} & 130   & 126   & \textbf{119} & 96    & \textbf{81} & \textbf{69} \bigstrut\\
    \hline
    \multicolumn{1}{|c|}{\multirow{4}[8]{*}{CAMI2\newline{}Urog}} & MaxBin2 & 37    & 36    & 36    & 35    & 34    & 29    & 26 \bigstrut\\
\cline{2-9}          & MetaBAT2 & 77    & 74    & 71    & 70    & 69    & 61    & 44 \bigstrut\\
\cline{2-9}          & VAMB  & 118   & 114   & 109   & 101   & 89    & 74    & 50 \bigstrut\\
\cline{2-9}          & CLMB  & \textbf{120} & \textbf{118} & \textbf{111} & \textbf{105} & \textbf{94} & \textbf{83} & \textbf{56} \bigstrut\\
    \hline
    \end{tabular}%
  \label{tab:st}%
\end{table}%

\begin{table}[tbp]
  \centering
  \caption{Number of genomes at the species level reconstructed with a precision of at least 95\%}
  
    \begin{tabular}{|c|c|c|c|c|c|c|c|c|}
    \hline
    \multicolumn{2}{|c|}{} & \multicolumn{7}{p{20.03em}|}{RECALL} \bigstrut\\
    \hline
    \multicolumn{1}{|p{6em}|}{Dataset} & \multicolumn{1}{p{6.68em}|}{Binner} & 0.50  & 0.60  & 0.70  & 0.80  & 0.90  & 0.95  & 0.99 \bigstrut\\
    \hline
    \multicolumn{1}{|c|}{\multirow{4}[8]{*}{CAMI2\newline{}Airways}} & \multicolumn{1}{p{6.68em}|}{MaxBin2} & 41    & 38    & 37    & 32    & 22    & 16    & 12 \bigstrut\\
\cline{2-9}          & \multicolumn{1}{p{6.68em}|}{MetaBAT2} & 76    & 69    & 63    & 53    & 38    & 28    & 17 \bigstrut\\
\cline{2-9}          & \multicolumn{1}{p{6.68em}|}{VAMB} & \textbf{98} & \textbf{97} & \textbf{95} & \textbf{90} & 61    & 45    & 27 \bigstrut\\
\cline{2-9}          & \multicolumn{1}{p{6.68em}|}{CLMB} & 95    & 92    & 91    & 85    & \textbf{66} & \textbf{47} & \textbf{30} \bigstrut\\
    \hline
    \multicolumn{1}{|c|}{\multirow{4}[8]{*}{CAMI2 GI}} & \multicolumn{1}{p{6.68em}|}{MaxBin2} & 59    & 58    & 58    & 55    & 51    & 48    & 44 \bigstrut\\
\cline{2-9}          & \multicolumn{1}{p{6.68em}|}{MetaBAT2} & 91    & 89    & 87    & 81    & 74    & 66    & 57 \bigstrut\\
\cline{2-9}          & \multicolumn{1}{p{6.68em}|}{VAMB} & 89    & 88    & 88    & 85    & 80    & 74    & 63 \bigstrut\\
\cline{2-9}          & \multicolumn{1}{p{6.68em}|}{CLMB} & \textbf{101} & \textbf{100} & \textbf{99} & \textbf{96} & \textbf{92} & \textbf{85} & \textbf{71} \bigstrut\\
    \hline
    \multicolumn{1}{|c|}{\multirow{4}[8]{*}{CAMI2\newline{}Oral}} & \multicolumn{1}{p{6.68em}|}{MaxBin2} & 63    & 60    & 54    & 46    & 39    & 31    & 21 \bigstrut\\
\cline{2-9}          & \multicolumn{1}{p{6.68em}|}{MetaBAT2} & 87    & 85    & 83    & 78    & 72    & 57    & 38 \bigstrut\\
\cline{2-9}          & \multicolumn{1}{p{6.68em}|}{VAMB} & \textbf{129} & \textbf{126} & \textbf{124} & \textbf{116} & \textbf{103} & \textbf{84} & 58 \bigstrut\\
\cline{2-9}          & \multicolumn{1}{p{6.68em}|}{CLMB} & 123   & 122   & 119   & 111   & 101   & 83    & \textbf{59} \bigstrut\\
    \hline
    \multicolumn{1}{|c|}{\multirow{4}[8]{*}{CAMI2\newline{}Skin}} & \multicolumn{1}{p{6.68em}|}{MaxBin2} & 56    & 53    & 50    & 46    & 34    & 30    & 27 \bigstrut\\
\cline{2-9}          & \multicolumn{1}{p{6.68em}|}{MetaBAT2} & 100   & 92    & 88    & 73    & 63    & 52    & 42 \bigstrut\\
\cline{2-9}          & \multicolumn{1}{p{6.68em}|}{VAMB} & 107   & \textbf{103} & \textbf{100} & 87    & 69    & 59    & 48 \bigstrut\\
\cline{2-9}          & \multicolumn{1}{p{6.68em}|}{CLMB} & \textbf{108} & 101   & 99    & \textbf{94} & \textbf{75} & \textbf{64} & \textbf{56} \bigstrut\\
    \hline
    \multicolumn{1}{|c|}{\multirow{4}[8]{*}{CAMI2\newline{}Urog}} & \multicolumn{1}{p{6.68em}|}{MaxBin2} & 34    & 33    & 33    & 32    & 31    & 26    & 24 \bigstrut\\
\cline{2-9}          & \multicolumn{1}{p{6.68em}|}{MetaBAT2} & 66    & 64    & 62    & 61    & 60    & 54    & 39 \bigstrut\\
\cline{2-9}          & \multicolumn{1}{p{6.68em}|}{VAMB} & 69    & 69    & 67    & 64    & 59    & 53    & 39 \bigstrut\\
\cline{2-9}          & 
\multicolumn{1}{p{6.68em}|}{CLMB}  & \textbf{74} & \textbf{74} & \textbf{71} & \textbf{68} & \textbf{64} & \textbf{60} & \textbf{43} \bigstrut\\
    \hline
    \end{tabular}%
  \label{tab:s}%
\end{table}%

\begin{table}[tbp]
  \centering
  \caption{Number of genomes at the genus level reconstructed with a precision of at least 95\%}
    \begin{tabular}{|c|p{6.68em}|c|c|c|c|c|c|c|}
    \hline
    \multicolumn{2}{|c|}{} & \multicolumn{7}{p{20.075em}|}{RECALL} \bigstrut\\
    \hline
    \multicolumn{1}{|p{6em}|}{Dataset} & Binner & 0.50  & 0.60  & 0.70  & 0.80  & 0.90  & 0.95  & 0.99 \bigstrut\\
    \hline
    \multicolumn{1}{|c|}{\multirow{4}[8]{*}{CAMI2\newline{}Airways}} & MaxBin2 & 30    & 28    & 27    & 23    & 16    & 11    & 9 \bigstrut\\
\cline{2-9}          & MetaBAT2 & 48    & 42    & 38    & 31    & 23    & 16    & 9 \bigstrut\\
\cline{2-9}          & VAMB  & \textbf{52} & \textbf{51} & \textbf{50} & \textbf{49} & 33    & 19    & 8 \bigstrut\\
\cline{2-9}          & CLMB  & 51    & 50    & 49    & 46    & \textbf{36} & \textbf{23} & \textbf{12} \bigstrut\\
    \hline
    \multicolumn{1}{|c|}{\multirow{4}[8]{*}{CAMI2 GI}} & MaxBin2 & 38    & 37    & 37    & 35    & 32    & 31    & 29 \bigstrut\\
\cline{2-9}          & MetaBAT2 & \textbf{56} & \textbf{54} & \textbf{53} & 48    & 42    & 37    & 34 \bigstrut\\
\cline{2-9}          & VAMB  & 47    & 46    & 46    & 45    & 43    & 38    & 34 \bigstrut\\
\cline{2-9}          & CLMB  & 50    & 50    & 50    & \textbf{49} & \textbf{46} & \textbf{43} & \textbf{40} \bigstrut\\
    \hline
    \multicolumn{1}{|c|}{\multirow{4}[8]{*}{CAMI2\newline{}Oral}} & MaxBin2 & 42    & 41    & 40    & 37    & 32    & 25    & 18 \bigstrut\\
\cline{2-9}          & MetaBAT2 & 55    & 54    & 52    & 50    & 47    & 41    & 28 \bigstrut\\
\cline{2-9}          & VAMB  & \textbf{66} & \textbf{63} & \textbf{63} & \textbf{61} & \textbf{54} & \textbf{47} & 34 \bigstrut\\
\cline{2-9}          & CLMB  & 62    & 62    & 61    & 59    & 53    & 45    & \textbf{37} \bigstrut\\
    \hline
    \multicolumn{1}{|c|}{\multirow{4}[8]{*}{CAMI2\newline{}Skin}} & MaxBin2 & 46    & 44    & 41    & 38    & 30    & 27    & 24 \bigstrut\\
\cline{2-9}          & MetaBAT2 & \textbf{64} & \textbf{61} & \textbf{61} & 52    & 46    & 39    & 33 \bigstrut\\
\cline{2-9}          & VAMB  & 58    & 58    & 56    & 50    & 44    & 37    & 31 \bigstrut\\
\cline{2-9}          & CLMB  & 57    & 55    & 55    & \textbf{54} & \textbf{48} & \textbf{41} & \textbf{36} \bigstrut\\
    \hline
    \multicolumn{1}{|c|}{\multirow{4}[8]{*}{CAMI2\newline{}Urog}} & MaxBin2 & 28    & 28    & 28    & 27    & 26    & 23    & 21 \bigstrut\\
\cline{2-9}          & MetaBAT2 & \textbf{35} & \textbf{34} & \textbf{33} & \textbf{32} & \textbf{32} & \textbf{29} & \textbf{23} \bigstrut\\
\cline{2-9}          & VAMB  & 29    & 29    & 29    & 26    & 24    & 22    & 18 \bigstrut\\
\cline{2-9}          & CLMB  & 33    & 33    & 31    & 30    & 29    & 27    & 21 \bigstrut\\
    \hline
    \end{tabular}%
  \label{tab:g}%
\end{table}%

\newpage

\section{Methods}
In this section, we show the methods and experiments in our research.

\subsection{Feature Calculation of TNFs and Abundance}
\label{m:f}
We use the same approach to calculate TNFs and abundance as the previous work \cite{nja21}.
For each contig, we count the frequencies of each tetramer with definite bases, and, to satisfy statistical constraints, project them into a 103-dimensional independent orthonormal space to obtain TNFs \cite{kbd09}.
As a result, the TNFs for each contig are a 103-dimensional numerical vector.
We also count the number of individual reads mapped to each contig.
More specifically, a read mapped to $n$ contigs counts $1/n$ towards each.
The read counts are normalized by sequence length and total number of mapped reads, which generates the abundance value in reads per kilobase sequence per million mapped reads (RPKM).
The resulted abundance for each contig is a $s$-dimensional numerical vector, where $s$ is the number of samples.
TNFs are normalized by z-scaling each tetranucleotide across the sequences, and abundance are normalized across samples.


\subsection{Benchmarking}
\label{m:b}
CLMB and VAMB \cite{nja21} were run with default parameters with multi-split enabled. MetaBAT2 \cite{klkteaw19} was run with setting minClsSize=1 and other parameters as default. MaxBin2 \cite{wss16} was run with default parameters. The benchmarking results were calculated using benchmark.py script implemented by \cite{nja21}. The mapping of the recovered genomes to the reference genomes was the intermediate result\footnote{The variable \textbf{recprecof} in function \textbf{Binning.\_getcounts()}} of benchmark.py script. FastANI \cite{jrp18} with default parameters was used to calculate ANI between the reference genomes. For the binning refinement experiment, we use metaWRAP bin\_refinement API \cite{udt18,sb73-5} with parameters –c 50 and –x 10, indicating we keep the genomes qualifying $completeness>50\%$ and $contamination<10\%$. The completeness and contamination of the genomes recovered by the bins are calculated using CheckM \cite{pisht15} with default parameters. We use the pipeline  integrated in MetaGEM \cite{zbpz21} for binning refinement experiment.

\subsection{Data fusion experiment}
\label{m:d}
We define the feature data as the raw data, and obtained the projected data by projecting the feature data to 32-dimension space using PCA. For the CLMB-encoded data, we obtained them by encoding the feature data to 32-dimension space with the deep contrastive learning framework. We assess the performance of these data by clustering them with the iterative medoid clustering and obtained the benchmarking results. All the experiments on CAMI2 datasets were run with default parameters with multi-split enabled, and the experiments on MetaHIT datasets was run with default parameters with multi-split disabled. For comparison to other clustering methods, we use MiniBatchKMeans (n\_clusters=750, batch\_size=4096, max\_iter=25, init\_size=20000, reassignment\_ratio=0.02) and DBSCAN (eps=0.35, min\_samples=2) implemented by scikit-learn.

\subsection{Binning of the Mother-Infant Transmission Dataset}
\label{m:m}
We downloaded the sequencing datasets of selected mother-infant pairs (marked as 10001, 10002, 10003, 10005, 10006, 10007, 10008, 10009, 10015, 10019) using SRA Toolkit and filtered them based on quality using fastp \cite{czcf18}.
Then, we assembled the short sequence reads into contigs using MEGAHIT \cite{lllsl15,lllltsyl16} and mapped the reads to the contigs using kallisto \cite{bpmp16} in order to speed up this process for large datasets.
The coabundance across samples can be subsequently calculated using kallisto quantification algorithm. With the assemblies and coabundances, we ran CLMB with default parameters and multi-split enabled.
Then, we splited the fasta file into bins based on the result of clustering using create\_fasta.py script.
CheckM \cite{pisht15} on lineage specific workflow with default parameters was applied to the resulting bins to calculate the completeness and contamination, and only those with sufficient quality ($completeness\geq 50\%$, $contamination\leq 5\%$) were considered for further analysis. Then, we use GTDB-tk \cite{cmhp20} on for taxonomic assignment of each bins and phylogeny inference. We visualized the tree with iTOL \cite{lb21}.

\newpage

\bibliographystyle{myrecomb}

\end{document}